\documentclass{article}
\usepackage{spconf,amsmath,graphicx}
\usepackage{bm}
\usepackage{amsfonts}       
\usepackage{amsmath}
\usepackage{graphicx}
\usepackage{subfigure}
\usepackage{booktabs}       

\title{Residual Encoder-Decoder Network for Deep Subspace Clustering}
\name{Shuai Yang, Wenqi Zhu, Yuesheng Zhu\thanks{This work was supported in part by the Shenzhen Municipal Development and Reform Commission (Disciplinary Development Program for Data Science and Intelligent Computing), in part by Shenzhen International cooperative research projects GJHZ20170313150021171, and in part by NSFC-Shenzhen Robot Jointed Founding (U1613215).
}}
\address{Institute of Big Data Technologies, Shenzhen Graduate School, Peking University, China\\}
\begin{document}
%
\maketitle
\begin{abstract}
  Subspace clustering aims to cluster unlabeled data that lies in a union of low-dimensional linear subspaces. Deep subspace clustering approaches based on auto-encoders have become very popular to solve subspace clustering problems. However, the training of current deep methods converges slowly, which is much less efficient than traditional approaches. We propose a Residual Encoder-Decoder network for deep Subspace Clustering (RED-SC), which symmetrically links convolutional and deconvolutional layers with skip-layer connections, with which the training converges much faster. We use a self-expressive layer to generate more accurate linear representation coefficients through different latent representations from multiple latent spaces. Experiments show the superiority of RED-SC in training efficiency and clustering accuracy. Moreover, we are the first one to apply residual encoder-decoder on unsupervised learning tasks.
\end{abstract}
\section{Introduction}
\label{sec:intro}
In this paper, we tackle the problem of Subspace Clustering (SC), which is a sub-field of unsupervised learning, aiming to cluster data points drawn from a union of low-dimensional subspaces in an unsupervised manner. Suppose that $X = \left[ \bm{x}_1,\dots,\bm{x}_N \right]\in \mathbb{R}^{D\times N}$ represents data set with $N$ data points in ambient dimension $D$, and data points lie in $n$ subspaces $\left\{S_i\right\}_{i=1}^{n}$ of dimensions $\left\{d_i\right\}_{i=1}^{n}$ ($d_i \ll \operatorname{min}\left\{D,N\right\}$). The task of SC is to partition data points into clusters $\left\{A_i\right\}_{i=1}^{n}$ so that data points within the same cluster $A_i$ lie in the same intrinsic subspace $S_i$. SC has achieved great success in many applications, e.g., motion segmentation \cite{2}, face clustering \cite{3} and image representation and compression \cite{4}.

Most traditional SC algorithms \cite{6, 7, 8, 9, 10, 16} are based on the linear subspace assumption to construct the affinity matrix for spectral clustering. However, the data doesn't necessarily conform to a linear subspace model, which motivates non-linear SC techniques. Kernel methods \cite{12, 13, 14} can be employed to implicitly map data to higher dimensional spaces for better conforming to linear models in the resulting spaces. However, the selection of different kernel types is largely empirical without theoretical guarantee. Recently, Convolutional Neural Networks has shown the superior ability in learning image representation, and Deep Subspace Clustering Networks (DSC-Net) \cite{15} have been proposed to exploit the self-expression of data in a union of subspaces.

Despite the significant improvements of clustering accuracy, DSC-Net suffers from the slow training compared with conventional ``shallow" SC methods. To achieve higher model training efficiency and higher clustering accuracy, we propose a Residual Encoder-Decoder network for deep Subspace Clustering (RED-SC). In particular, we make the following contributions:
\begin{itemize}
  \item We propose to establish skip connections between corresponding convolutional and deconvolutional layers. These skip connections help to back-propagate the gradients to bottom layers and pass data details to top layers, making training of the end-to-end mapping easier and more effective.
  \item We propose to insert the self-expressive layer in each skip connection to generate the linear representation coefficients. We present a new global loss function and minimize it by RED-SC. This helps to learn the linearity information of features in different latent spaces.
  \item To the best of our knowledge, our approach constitutes the first attempt to apply residual encoder-decoder network on the task of unsupervised learning.
\end{itemize}

Experimental results demonstrate that our network converges much faster in model training and fine-tuning, and obtains better clustering results. We reduce the computational cost remarkably, and obtain higher accuracy simultaneously.

\begin{figure*}
  \label{fig:1}
  \centering
  \resizebox{.4\width}!{
  \includegraphics{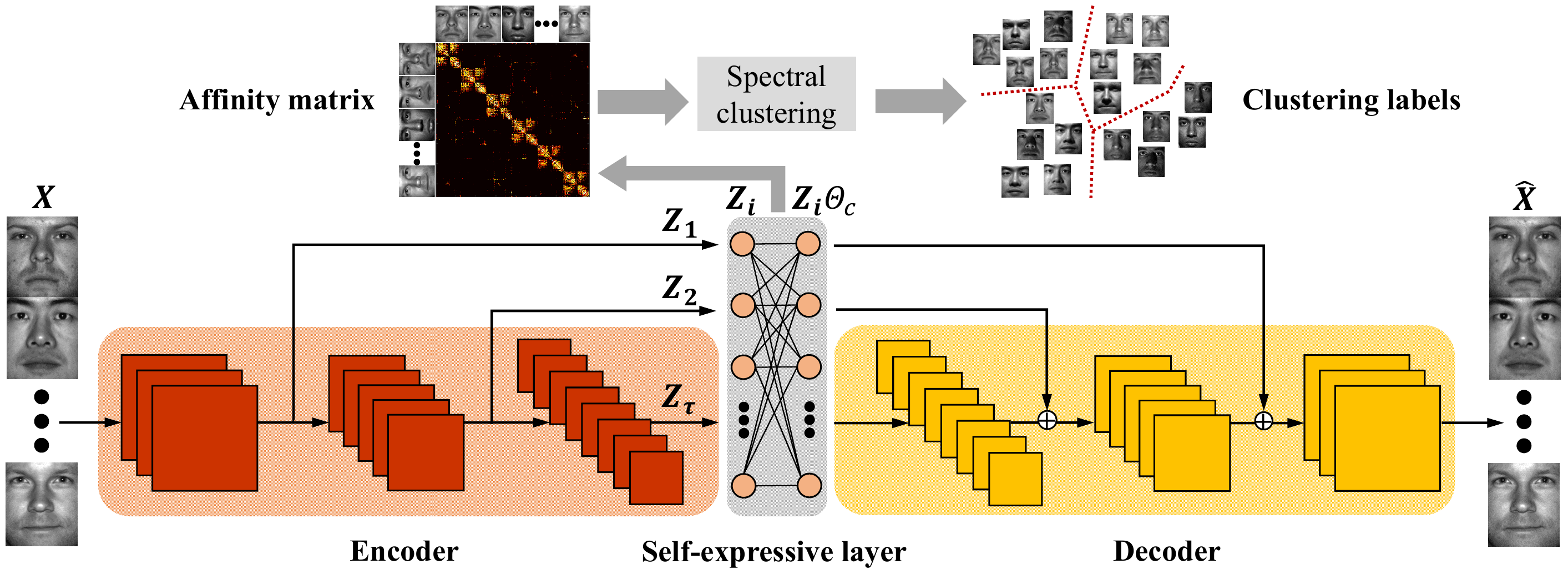}}
  \caption{The proposed Residual Encoder-Decoder network for deep Subspace Clustering (RED-SC).}
\end{figure*}

\section{Related Work}
\label{sec:related work}
\subsection{Subspace Clustering}
Many methods have been developed for linear subspace clustering. Generally, these approaches are based on a two-stage framework. In the first stage, an affinity matrix is generated from data by computing the linear representation coefficients matrix $C$. In the second one, spectral clustering is applied on the affinity matrix. These methods learn the affinity matrix based on the self-expressiveness model, which states that each data point in a union of subspaces can be expressed as a linear combination of other data points, i.e., $X = XC,$ where $X$ is the data matrix, $C\in \mathbb{R}^{N\times N}$ is the coefficients matrix. To find the coefficients matrix $C$, current methods solve the following optimization problem in the first stage:
\begin{equation}
\label{equ:1}
\min _{C}\|C\|_{p},\quad \text { s.t.} \quad X=XC, \operatorname{diag}(C)= 0,
\end{equation}
where $\|\cdot\|_{p}$ denotes different norm regularization applied on $C$. For instance, in Sparse Subspace Clustering (SSC) \cite{6}, the $\ell_1$ norm regularization is adopted as a convex surrogate over the $\ell_0$ norm regularization to encourage the sparsity of $C$. Least Squares Regression (LSR) \cite{7} uses the $\ell_2$ norm regularization on $C$. Low Rank Representation (LRR) \cite{8} uses nuclear norm regularization on $C$. Elastic Net Subspace Clustering (ENSC) \cite{9} uses a mixture of $\ell_1$ norm and $\ell_2$ norm regularization on $C$. In SSC by Orthogonal Matching Pursuit (OMP) \cite{10} and our previous work Sparse-Dense Subspace Clustering (SDSC) \cite{11}, the $\ell_0$ norm regularization is investigated. However, they can only cluster linear subspaces, which limits their application. To address this problem, kernel based subspace clustering methods \cite{12, 13, 14} have been developed. There is, however, no clear reason why such kernels should correspond to feature spaces that are well-suited to subspace clustering. Recently, Deep Subspace Clustering Networks (DSC-Net) \cite{15} are introduced to tackle the nonlinearity arising in subspace clustering, where data is nonlinearly mapped to a latent space with convolutional auto-encoders and a self-expressive layer is introduced to facilitate an end-to-end learning of the coefficients matrix. Although DSC-Net outperforms traditional SC methods, the computational cost especially in model training is overwhelming.
\subsection{Residual Encoder-Decoder}
Encoder-decoder networks can non-linearly map data into a latent space. It can be viewed as a form of non-linear PCA if the latent space has lower dimension than the original space \cite{18}. Residual encoder-decoder networks with skip-layer connections have been exploited effective in many applications, e.g., image restoration \cite{19}, semantic segmentation \cite{20} and iris segmentation \cite{21}. It has been shown that residual encoder-decoder networks converge much faster in model training since the skip connections help to back-propagate the gradients to bottom layers and pass image details to top layers, making training of the end-to-end mapping easier. Besides, the feature maps passed by skip
connections carry much image detail, which helps deconvolution to recover a better and cleaner image. To the best of our knowledge, it has not been used in any tasks of unsupervised learning. Our RED-SC to solve subspace clustering problems constitutes the first attempt to apply residual encoder-decoder on the tasks of unsupervised learning.
\section{Residual Encoder-Decoder Network for Deep Subspace Clustering (RED-SC)}
\label{sec:RED-SC}
The proposed network uses the residual encoder-decoder and the self-expressiveness property. In this section, we first discuss each component, then introduce the network architecture, and finally elaborate its training and clustering process.

\subsection{Residual Encoder-Decoder in RED-SC}
The DSC-Net uses auto-encoders to map the data into a latent space, then uses the feature in the latent space to generate the linear representation coefficients for affinity matrix, and finally recover the data by a chain of decoders. But the intuitive question is that, is deconvolution able to recover the data detail from the abstraction only? Another question is that, can the feature from only one latent space represent the data to generate the linear representation coefficients? We find that much data detail is lost in the convolution, making DSC-Net hard to train, and the affinity matrix is inaccurate to represent the relationship of original data.

\begin{figure*}
  \centering
  \resizebox{.66\width}!{
  \includegraphics{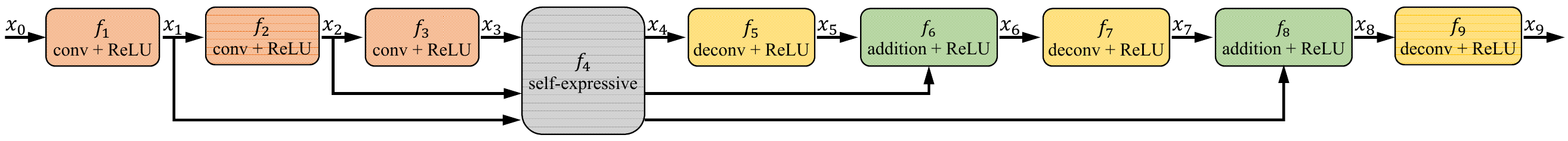}}
  \caption{An example of a building block in the proposed RED-SC.}
  \label{fig:2}
\end{figure*}

To address the above two problems, inspired by residual networks \cite{22} and highway networks \cite{23}, we add skip connections between two corresponding convolutional and deconvolutional layers as shown in Fig.\ref{fig:1}. A building block is shown in Fig.\ref{fig:2}. Instead of directly learning the mappings from input $X$ to the output $Y$, we would like the network to fit the residual of the problem, which is denoted as:
\begin{equation}
\label{equ:2}
\mathcal{F}(X)=Y-X.
\end{equation}
Such a learning strategy is applied on inner blocks of the encoding-decoding network to make training more effective.

By using the residual encoder-decoder network, the feature maps passed by skip connections carry much data detail, which helps deconvolution to better recover the data. Besides, the skip connections also achieve benefits on back-propagating the gradient to bottom layers, which avoids the network suffering from gradient vanishing.

\subsection{Self-Expressive Layer in RED-SC}
Recall from the optimization problem in (\ref{equ:1}), to account for data corruptions, this problem is relaxed as:
\begin{equation}
\min _{C}\|C\|_{p}+\frac{\lambda}{2}\|X-XC\|_{F}^{2} \quad \text {s.t.} \quad\operatorname{diag}(C)=0.
\end{equation}
In our RED-SC network, latent representation from multiple layers are adopted as input of the self-expressive layer to generate the self-expressive coefficients. Let $X \in \mathbb{R}^{D\times N}$ denote the input data, and $Z_i\in \mathbb{R}^{D\times N}$ denote the output of each convolution layer, we introduce a self-expression loss as:
\begin{equation}
\mathcal{L}_{s}\left(\Theta_{c} ; Z\right)=\sum_{i=1}^{\tau}\left\|Z_i-Z_i \Theta_{c}\right\|_{F}^{2}+\lambda\left\|\Theta_{c}\right\|_{F}^{2},
\end{equation}
where $\tau$ is the number of convolutional layer,  $\Theta_{c} \in \mathbb{R}^{N\times N}$ is the self-expressive coefficients matrix.
Our goal is to train a deep residual encoder-decoder network, we can calculate the reconstruction loss of data $X$ after the network as:
\begin{equation}
\mathcal{L}_{e}\left(\Theta_{c}, \Theta_{e}, \Theta_{d} ; X\right)=\frac{1}{2}\left\|X - \hat{X}\right\|_{F}^{2},
\end{equation}
where $\hat{X}$ represents the data reconstructed by the residual encoder-decoder, $\Theta_{e}$ and $\Theta_{d}$ respectively represent the encoder parameters and the decoder parameters. Then we can compute the global loss of our RED-SC network:
\begin{equation}
\label{equ:6}
\begin{aligned}
&\mathcal{L}\left(\Theta\right) = \mathcal{L}_{s}\left(\Theta_{c} ; Z\right) + \mathcal{L}_{e}\left(\Theta_{c}, \Theta_{e}, \Theta_{d} ; X\right)\\
&= \frac{1}{2}\left\|X - \hat{X}\right\|_{F}^{2} + \sum_{i=1}^{\tau}\left\|Z_i-Z_i \Theta_{c}\right\|_{F}^{2}+\lambda\left\|\Theta_{c}\right\|_{F}^{2},
\end{aligned}
\end{equation}
where the network parameters $\Theta$ consist of $\Theta_{c}$, $\Theta_{e}$ and $\Theta_{d}$. In this work, we consider the $\ell_2$ norm regularization on $\Theta_c$ for computational efficiency.
\subsection{Network Architecture}
In this paper, we focus on image clustering problems. As is shown in Fig.\ref{fig:1}, we use all the images as a single batch. The input images $X$ are mapped to a collection of latent vectors $Z_i$ by each convolutional layer. In the self-expressive layer, the nodes are fully connected using linear weights without bias and non-linear activations. Then the latent vectors of skip connections are mapped into symmetric layer in decoder for addition and non-linear activations (ReLU) \cite{24}, and the output of the last convolutional layer is mapped back into the original space by deconvolutional layers in decoder. Finally we use the self-expressive coefficients to generate the affinity matrix, and then apply spectral clustering on the affinity matrix to get the clustering labels.

In particular, for the $i$th convolutional layer with $n_i$ channels of kernel size $k_i \times k_i$, there are $k^2_{i}n_{i-1}n_i$ weight parameters. The total number of weight parameters in our network is $\sum_{i=1}^{\tau}2k^2_{i}n_{i-1}n_i$, and that of bias parameters is $\sum_{i=1}^{\tau}2n_i-n_1+1$. Suppose the number of input samples is $N$, then the number of self-expressive parameters is $N^2$, which is much larger than the number of weights and bias parameters. Thus the self-expressive parameters dominate the network.
\subsection{Training Strategy}
\begin{figure}
  \centering
  \resizebox{1\columnwidth}!{
  \includegraphics{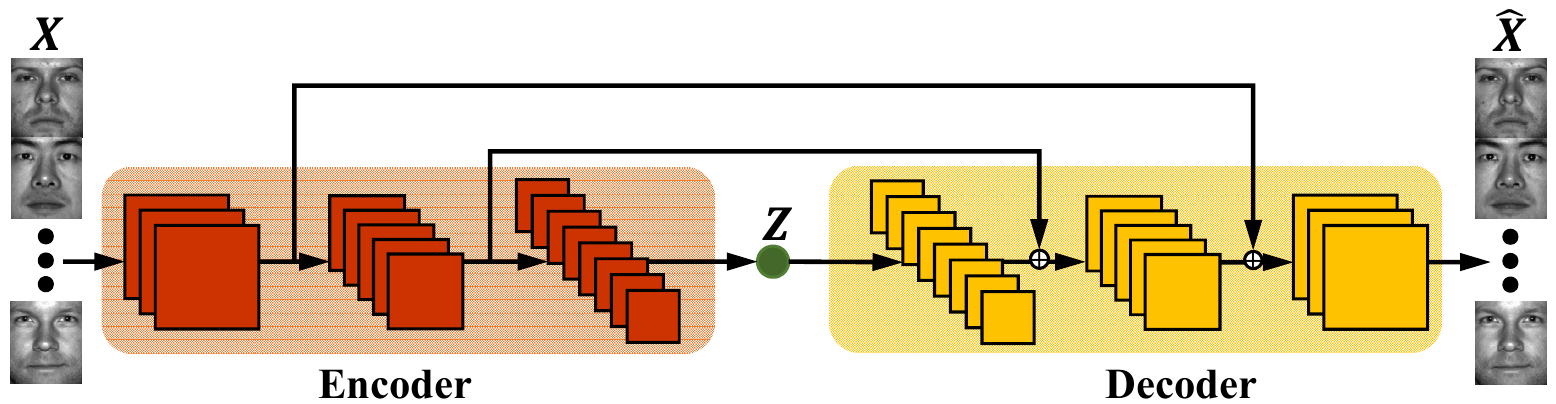}}
  \caption{The pre-training network.}
  \label{fig:3}
\end{figure}

Due to the limited size of data sets for unsupervised subspace clustering, it's difficult to train a network with millions of parameters. Thus we design a pre-training network without self-expressive layer in Fig.\ref{fig:3}. Then we use the trained parameters to initialize the encoder and decoder layers in our fine-tuning network with the self-expressive layer. With the help of Adam \cite{25}, we then use a big batch of all the data to minimize the loss $\mathcal{L}\left(\Theta\right)$ defined in (\ref{equ:6}). Note that we don't use any label information to train the model, our training strategy remains unsupervised. Finally, we use the trained self-expressive coefficients to construct the affinity matrix for spectral clustering, and get the clustering labels.

\section{Experiments}
\label{sec:exp}
We implement our approach with Tensorflow \cite{26} on a NVIDIA TITAN Xp GPU, and evaluate the performance of RED-SC on a handwritten digit data set MNIST \cite{27}, and a face data set Extended Yale B \cite{28}. We compare our RED-SC with LRR \cite{8}, LRSC \cite{16}, SSC \cite{6}, SSC-OMP \cite{10}, SSSC \cite{17}, SDSC \cite{11}, EDSC \cite{9} and DSC-Net \cite{15} with two norm regularization. We use the code provided by the respective authors which is tuned to give the best performance. We evaluate the clustering performance by using clustering error (ERR) \cite{15}, normalized mutual information (NMI) \cite{29} and purity (PUR) \cite{30}. For RED-SC, the kernel sizes are always 5-3-3-3-3-5 and channels are 10-20-30-30-20-10.
We use the pre-training network to obtain the parameters for the fine-tuning networks. The best results in tables are in bold.
\subsection{Experiments on MNIST}
We evaluate the effectiveness of RED-SC on MNIST, which consists of 70,000 hand-written digit images of size $28 \times 28$. We randomly select 1,000 images for each digit, resulting a subset of 10,000 images. For traditional SC algorithms LRR, SSC and ENSC, we use a subset of 1,000 images due to their limited scalability. The results are reported in Table 1.
\begin{table}[htbp]
  \centering
  \caption{Performance on MNIST.}
  \small
    \begin{tabular}{c|c|c|c}
    \toprule
          & ERR (\%) & NMI (\%) & PUR (\%) \\
    \hline
    LRR   & 46.25 & 56.32 & 56.84 \\
    SSC   & 55.71 & 47.09 & 49.41 \\
    ENSC  & 50.17 & 54.94 & 54.83 \\
    DSC-$\ell_1$ & 32.22 & 67.17 & 73.87 \\
    DSC-$\ell_2$ & 30.09 & 68.64 & 74.31 \\
    \textbf{RED-SC} & \textbf{25.66} & \textbf{73.16} & \textbf{77.24} \\
    \bottomrule
    \end{tabular}%
  \label{tab:1}%
\end{table}%

We can observe that RED-SC outperforms the traditional SC algorithms greatly, this is partly because RED-SC uses a multi-layer encoder as the feature extractor. Besides, compared with the deep approach DSC-Net, our RED-SC obtains better performance in all three metrics. This is because RED-SC tunes the self-expressive coefficients in multiple latent spaces, while DSC-Net only uses the latent representation from the last convolutional layer. This experimental result demonstrates the effectiveness of RED-SC to ensure better self-expressive coefficients for spectral clustering.
\subsection{Experiments on Extended Yale B}
We evaluate the efficiency of RED-SC in model training and fine-tuning on Extended Yale B, which contains 2,414 frontal face images of 38 individuals under 9 poses and 64 illumination conditions. Each cropped face image consists of 192$\times$168 pixels. We downsample the images to 48$\times$42 pixels. We randomly pick $n \in \left\{10, 15, 20, 25, 30, 38\right\}$ subjects and take all the images of selected subjects to be clustered.
\begin{table}[ht]
  \centering
  \small
  \caption{Average ERR (\%) on Extended Yale B.}
    \begin{tabular}{c|cccccc}
    \toprule
    Subjects & 10    & 15    & 20    & 25    & 30    & 38 \\
    \hline
    LRR   & 19.76 & 25.82 & 31.45 & 28.14 & 38.59 & 35.12 \\
    LRSC  & 30.95 & 31.47 & 28.76 & 27.81 & 30.64 & 29.89 \\
    SSC   & 8.8   & 12.89 & 20.11 & 26.3  & 27.52 & 29.36 \\
    SSC-OMP & 12.08 & 14.05 & 15.16 & 18.89 & 20.75 & 23.52 \\
    SSSC  & 6.34  & 11.01 & 14.07 & 16.79 & 20.46 & 19.45 \\
    SDSC  & 4.62  & 8.31  & 11.87 & 14.55 & 16.87 & 16.17 \\
    EDSC  & 5.64  & 7.63  & 9.3   & 10.67 & 11.24 & 11.64 \\
    DSC-$\ell_1$ & 2.23  & 2.17  & 2.17  & 2.53  & 2.63  & 3.33 \\
    DSC-$\ell_2$ & 1.59  & 1.69  & 1.73  & 1.75  & 2.07  & 2.67 \\
    \textbf{RED-SC} & \textbf{1.25} & \textbf{1.30} & \textbf{1.37} & \textbf{1.42} & \textbf{1.45} & \textbf{1.48} \\
    \bottomrule
    \end{tabular}%
  \label{tab:2}%
\end{table}%

\begin{figure}[ht]
\label{fig:4}
  \centering
  \subfigure[]{
  \includegraphics[width = 0.47\columnwidth]{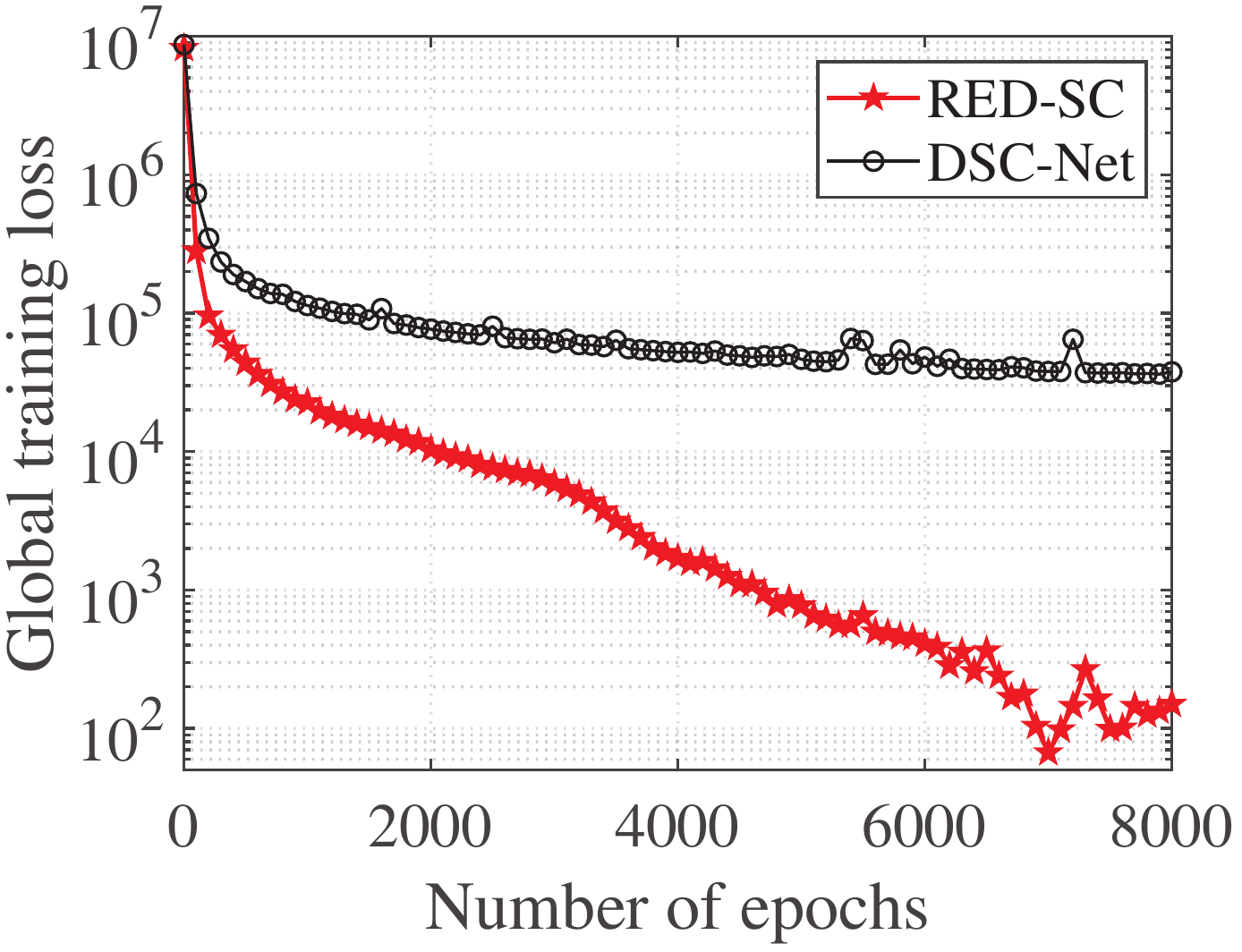}
  \label{fig:4a}
  }
  \subfigure[]{
  \includegraphics[width = 0.47\columnwidth]{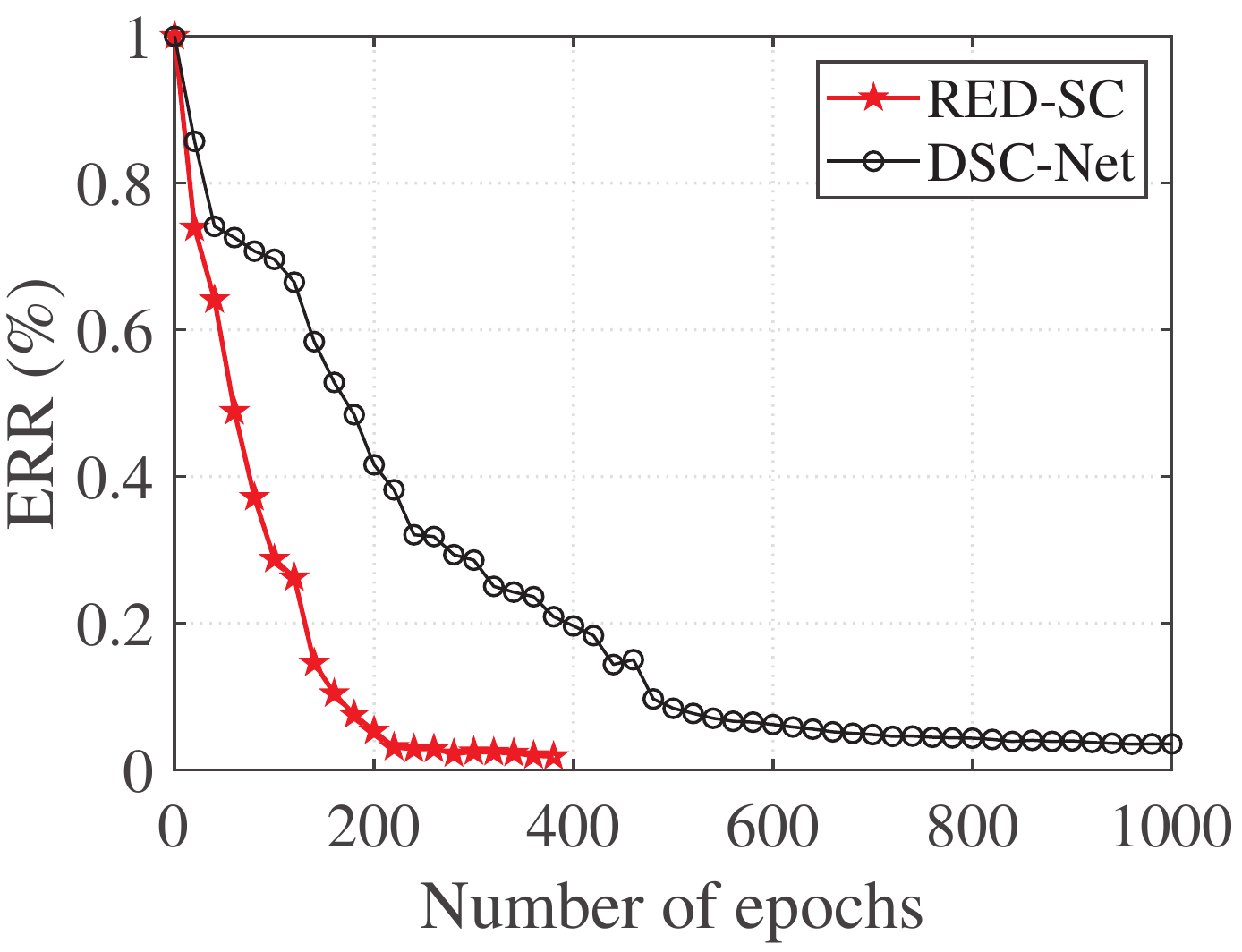}
  \label{fig:4b}
  }
  \caption{(a) The convergence of the global loss in pre-training network; (b) The convergence of ERR in fine-tuning network of the 38 subjects clustering problem on Extended Yale B.}
\end{figure}

As is shown in Table \ref{tab:2}, RED-SC remarkably reduces the clustering error and outperforms all the listed methods. This demonstrates again the effectiveness of RED-SC. Besides, we report the convergence compared with DSC-Net with the same number of parameters in Fig.4. From Fig.4(a) we observe that RED-SC converges much faster than DSC-Net in training, since the residual encoder-decoder architecture helps back-propagate gradient to better fit the end-to-end mapping. From Fig.4(b) we observe that RED-SC generates a high-quality affinity matrix for spectral clustering by approximately 300 epoches, while DSC-Net uses about 1,000 epoches. This is partly because that RED-SC uses the latent representation from multiple convolutional layers to fine-tune the self-expressive coefficients, which accelerates the convergence. Thus RED-SC gains a higher efficiency.
\section{Conclusion}
\label{sec:cncus}
We present a Residual Encoder-Decoder network for deep Subspace Clustering (RED-SC), which symmetrically links convolutional and deconvolutional layers with skip-layer connections. We present a new global loss and minimize it by RED-SC. We are the first one to apply residual encoder-decoder on unsupervised learning tasks. Series of experiments validate that RED-SC remarkably reduces computational cost and improves clustering performance.

\bibliographystyle{IEEEbib}
\bibliography{refs}
\end{document}